\pdfoutput=1

\documentclass[11pt]{article}
\usepackage{graphicx} 
\usepackage[table]{xcolor}
\definecolor{mygray}{gray}{0.9}
\usepackage[]{acl}
\usepackage{booktabs}
\usepackage{times}
\usepackage{amsmath}
\usepackage{amssymb}
\usepackage{latexsym}
\usepackage{graphicx}   
\usepackage{makecell}   
\usepackage{algorithm}
\usepackage{algpseudocode}
\usepackage{amsmath}
\usepackage[T1]{fontenc}
\usepackage{multirow} 
\usepackage[utf8]{inputenc}
\usepackage{mdframed}
\usepackage{lipsum}  

\usepackage{microtype}

\usepackage{inconsolata}
\usepackage[absolute,overlay]{textpos}
\usepackage{float}

\title{CounterBench: Evaluating and Improving
Counterfactual Reasoning in Large Language
Models}

\author{
Yuefei Chen$^1$\quad
Vivek K.Singh$^1$\quad 
Jing Ma$^2$\quad 
\textbf{Ruixiang Tang}$^{1}$\thanks{Corresponding Email: ruixiang.tang@rutgers.edu }\\
  $^1$Rutgers University\;\;\;  $^2$Case Western Reserve University\;\;\;
}
 
\begin{document}
\maketitle
\begin{abstract}
Counterfactual reasoning is widely recognized as one of the most challenging and intricate aspects of causality in artificial intelligence. In this paper, we evaluate the performance of large language models (LLMs) in counterfactual reasoning. In contrast to previous studies that primarily focus on commonsense causal reasoning, where LLMs often rely on prior knowledge for inference, we specifically assess their ability to perform counterfactual inference using a set of formal rules.  To support this evaluation, we introduce a new benchmark dataset, \textbf{CounterBench}, comprising 1.2K counterfactual reasoning questions. The dataset is designed with varying levels of difficulty, diverse causal graph structures, distinct types of counterfactual questions, and multiple nonsensical name variants. Our experiments demonstrate that counterfactual reasoning poses a significant challenge for LLMs, with most models performing at levels comparable to random guessing. To enhance LLM's counterfactual reasoning ability, we propose a novel reasoning paradigm, \textbf{CoIn}, which guides LLMs through iterative reasoning and backtracking to systematically explore counterfactual solutions. Experimental results show that our method significantly improves LLM performance on counterfactual reasoning tasks and consistently enhances performance across different LLMs. Our dataset is available at \href{https://huggingface.co/datasets/CounterBench/CounterBench}{https://huggingface.co/datasets/CounterBench\\/CounterBench}. 
\end{abstract}

\section{Introduction}
\begin{figure}[t]
  \centering
  \includegraphics[width=0.3\textwidth]{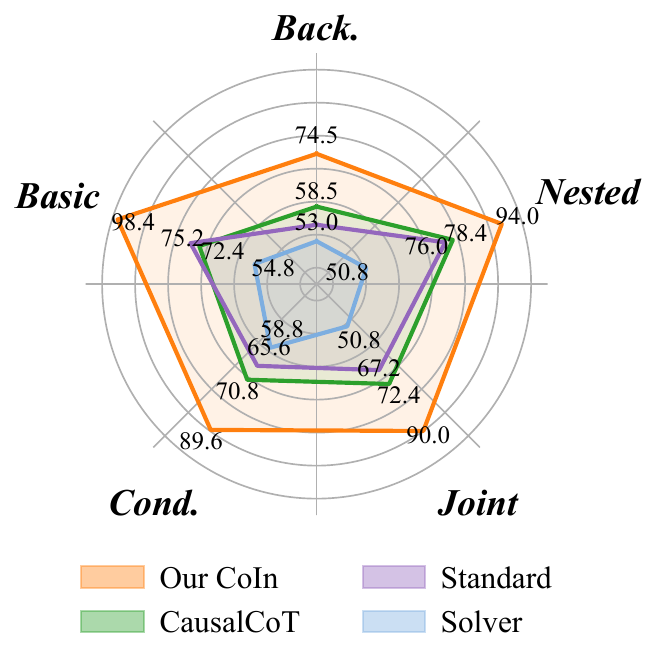}
  \caption{Comparison of accuracy scores on the CounterBench dataset across different strategies: our proposed CoIn paradigm versus baseline approaches (Standard, CausalCoT \cite{jin2023cladder}, and Solver \cite{hua2024improving}), evaluated using Gemini-1.5-flash. Our CounterBench dataset includes five kinds types. Basic focuses on exploring how a single change in a causal variable. Joint involves simultaneous changes in multiple causes, Nested involves stepwise hypothetical assumptions about multiple variables. Conditional evaluates counterfactuals under observed conditions. And Backdoor involves counterfactual reasoning in the presence of backdoor paths that create confounding between the treatment variable and the outcome.
  }
  \label{fig:fixed}
\end{figure}

Counterfactual reasoning, residing at the pinnacle of Pearl’s Causal Hierarchy \cite{pearl2018book}, underpins the “what if” inquiries essential to human cognition and decision-making across critical fields such as healthcare, business, public administration, and science \cite{gvozdenovic2021causal, kyrimi2025counterfactual, kasirzadeh2021use, koonce2011judging, gow2016causal, loi2012note}. For example, a consumer who declined an extended warranty may later wonder, ``What if I had purchased it, could I have avoided the repair costs?'' This illustrates how counterfactual reasoning guides decision-making by evaluating missed opportunities and alternative outcomes \cite{krishnamurthy2002counterfactual}. While traditional causal inference methods \cite{sharma2020dowhy, chen2020causalml, feder2022causal} have enhanced the predictive accuracy, robustness, and explainability of NLP models, recent progress in LLMs has further enriched our ability to capture nuanced causal dependencies \cite{liu2024proficient, petroni2019language, liang2024mapping, tarassow2023potential, ma2024causal, liu2024large}. These advancements not only demonstrate sophisticated reasoning in tasks ranging from writing to programming but also pave the way toward emulating human-like intelligence and achieving artificial general intelligence \cite{li2024relationfirstmodelingparadigmcausal, alwin2023causalai, sahota2023causalai, bubeck2023sparks}.

Despite recent advancements, progress in counterfactual reasoning using LLMs remains constrained by two primary challenges. First, there is currently no dedicated benchmark dataset for rigorously evaluating LLMs’ performance on counterfactual tasks, making it difficult to measure the models’ capacity to capture nuanced causal relationships. Second, even with advanced prompting techniques, Causal CoT \cite{jin2023cladder} and other iterative methods, LLMs often struggle to produce logically consistent, contextually appropriate counterfactuals \cite{ma2024causal,jin2023cladder,kiciman2023causal,zevcevic2023causal}. In response, this paper focuses on two key questions:

\textbf{How well do LLMs handle counterfactual reasoning?}
The absence of a standardized benchmark dataset has impeded rigorous empirical evaluation of LLMs' capabilities in capturing intricate causal relationships within complex counterfactual reasoning tasks. To address this, we present CounterBench, a comprehensive evaluation framework designed to assess counterfactual reasoning through 1.2K questions encompassing various domains and reasoning types. By systematically evaluating five key dimensions, it demands genuine reasoning beyond pattern recognition or memorized responses. Our experiments expose notable performance limitations in LLMs, even those equipped with advanced inference techniques. Most models like GPT-4o and Deepseek-V3 achieve accuracy of approximately $\textbf{50\%}$, equivalent to random guessing. Furthermore, our evaluation of state-of-the-art inference strategies shows only marginal improvements over baseline performance for most models. The models consistently struggle with maintaining logical coherence during multi-step reasoning processes and accurately handling causal relationships in complex scenarios. 

\textbf{How to improve LLMs’ counterfactual reasoning abilities?}
To advance large language models’ counterfactual reasoning capabilities, this paper presents CoIn (\textbf{Co}unterfactual \textbf{In}ference), a novel approach that explicitly tackles the critical challenges of multi-step inference, which remain unresolved by previous methods. CoIn embeds a tailored search algorithm into the reasoning process, guiding LLMs through abduction, action, iterative prediction, and backtracking validation to systematically formalize and explore counterfactual paths. This structured mechanism dynamically assesses the promise of each inference step, enabling reversion to more promising points and ensuring logical consistency, particularly in long-chain causal dependencies. This systematic process substantially improves the accuracy of counterfactual analysis.
Experiments on CounterBench demonstrate that CoIn achieves an accuracy of $\textbf{89.9\%}$, delivering a nearly $\textbf{20\%}$ improvement over Gemini-1.5-flash compared to alternative strategies (see Figure~\ref{fig:fixed}). The framework of this work is illustrated in Figure~\ref{fig:introduction}. The contributions of this work are summarized as follows:

\begin{figure*}[htbp]
  \centering
  \includegraphics[width=0.9\textwidth]{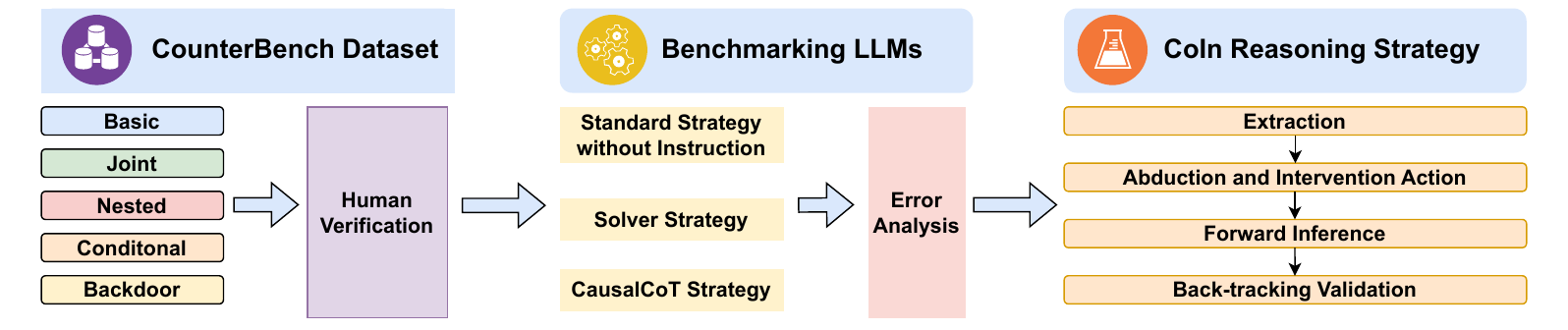}
  \caption{Illustration of the framework. We create CounterBench, a dataset featuring five types of counterfactual questions (basic, joint, conditional, nested, and backdoor). Based on this dataset, we benchmark state-of-the-art LLMs using various inference strategies, conduct comprehensive error analysis, and propose our CoIn reasoning framework featuring systematic inference with validation mechanisms.}
  \label{fig:introduction}
\end{figure*}

\begin{itemize}
    \item We build a comprehensive dataset, \textbf{CounterBench}. The dataset contains over 1200 long-chain complex counterfactual reasoning questions. The dataset spans multiple difficulty levels, diverse causal graph structures, various types of counterfactual questions, and a wide range of nonsensical variant name combinations. 
    \item We benchmark LLMs with various inference strategies on \textbf{CounterBench}, and results reveal that most existing models (e.g., GPT-4o and Deepseek-V3) exhibit limited capabilities in performing counterfactual inference tasks.
    \item We propose a novel reasoning paradigm \textbf{CoIn} guides LLMs through abduction, action, iterative prediction, and backtracking validation to systematically formalize and explore counterfactual reasoning paths. It achieves nearly 90\% accuracy on several state-of-the-art LLMs evaluated on CounterBench, representing a 20\% improvement over the previous best baseline.
    
\end{itemize}

\begin{table*}[t]
\centering

\setlength{\tabcolsep}{0.6mm}
\resizebox{\linewidth}{!}{
\begin{tabular}{l|l|m{5cm}}
\toprule
\rowcolor{mygray}
\textbf{Type} & \textbf{Query Template Example} & \textbf{Causal graph}\\ \hline

\textbf{Basic} & \makecell[l] {We know that X causes V1, V1 causes V2, V2 causes V3, and V3 causes V4, \\V4 causes V5, V5 causes Y. Would Y occur if not X instead of X?} &\includegraphics[width=5cm]{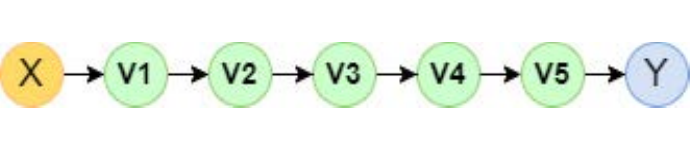}
 \\  \hline

\textbf{Joint} & \makecell[l]{We know that X causes V1, V1 causes V2, V2 and V1 together cause V3, \\V3 causes V4, V4 and X together cause V5, and V5 causes Y. \\Would Y occur if not X and not V3?} &\includegraphics[width=5cm]{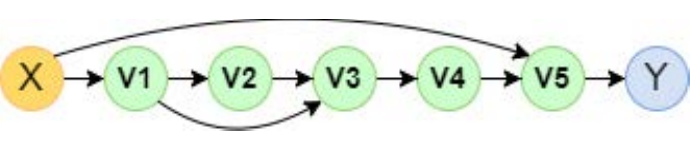}
\\ \hline 

\textbf{Nested} & \makecell[l]{We know that X causes V1, V1 causes V2, V2 and V1 together cause V3,\\ V3 causes V4, V4 and V2 together cause V5, and V5 causes Y.\\ Assume not X, and based on this assumption, further suppose not V4.\\ Would Y occur?} &\includegraphics[width=5cm]{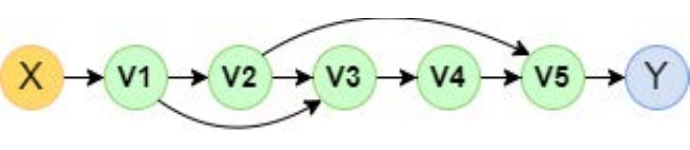}
\\ \hline

\textbf{Conditional} & \makecell[l]{We know that X and V1 together cause V2, V2 causes V3, V3 causes V4, \\V4 causes V5, V5 causes Y.  We observed V1. Would Y occur if not X \\instead of X?} & \includegraphics[width=5cm]{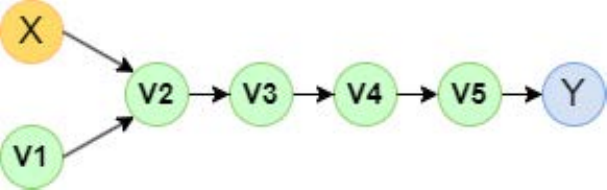}
\\ \hline

\textbf{Backdoor} & \makecell[l]{We know that V1 causes X, X and V1 together cause V2, V2 causes V3, \\V3 and X together cause V4, V4 causes V5, and V5 causes Y.} & \includegraphics[width=5cm]{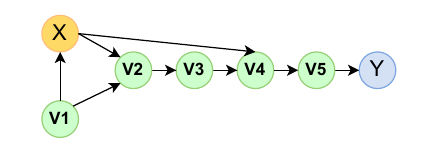}
\\ \hline
\end{tabular}
}
\caption{Illustrative Counterfactual Query Types}
\label{tab:counterfactual_types}
\end{table*}

\section{CounterBench}

To evaluate the counterfactual reasoning capabilities of LLMs, we introduce a comprehensive benchmarking dataset specifically designed to measure their ability to handle complex causal reasoning tasks. This section details the structure of the dataset, the methodology for query generation, and the benchmarking results analysis.

\subsection{Dataset Structure}

The dataset consists of two main components: a set of counterfactual queries and corresponding binary answers. Formally, the dataset is defined as $\mathcal{D} := \{(q_i, a_i) | i = 1, 2, ..., N\}$, where each $q_i$ is a counterfactual query, and $a_i \in \{\text{yes}, \text{no}\}$ represents the correct answer. Each query is derived from a deterministic Structural Causal Model (SCM) $M = \langle U, V, f \rangle$, where $U$ is the set of exogenous variables with assignments $u$, $V$ is the set of endogenous variables, and $f$ is the set of structural equations \cite{pearl2009causality}. For each $V_i \in V$, we have $V_i = f_i(Pa(V_i), U_i)$, where $Pa(V_i) \subseteq V$ denotes the parents of $V_i$, and $U_i$ refers to the subset of exogenous variables from $U$ that directly influence the value of $V_i$. Intervening on a set of variables $X \subseteq V$ and setting them to $x$ modifies the model to $M_x$, which deterministically defines the values of intervened variables given $u$. The dataset includes five types of counterfactual queries:

\begin{table*}[h!]
\centering
\setlength{\tabcolsep}{1.5mm}
\begin{tabular}{p{3.5cm}>{\centering\arraybackslash}p{0.8cm}>{\centering\arraybackslash}p{0.8cm}>{\centering\arraybackslash}p{0.8cm}>{\centering\arraybackslash}p{0.85cm}>{\centering\arraybackslash}p{0.8cm}>{\centering\arraybackslash}p{0.8cm}|>{\centering\arraybackslash}p{0.8cm}>{\centering\arraybackslash}p{0.8cm}>{\centering\arraybackslash}p{0.8cm}>{\centering\arraybackslash}p{0.85cm}>{\centering\arraybackslash}p{0.8cm}>{\centering\arraybackslash}p{0.8cm}}
\toprule
\rowcolor{mygray} \multicolumn{1}{c}{} & \multicolumn{6}{c}{\textbf{Standard}} & \multicolumn{6}{c}{\textbf{CausalCoT}} \\ \hline
\rowcolor{mygray} \textbf{Models} & \textbf{Basic} & \textbf{Cond.} & \textbf{Joint} & \textbf{Nested} & \textbf{Back.} & \textbf{Avg.} & \textbf{Basic} & \textbf{Cond.} & \textbf{Joint} & \textbf{Nested} & \textbf{Back.} & \textbf{Avg.} \\ \hline

GPT-3 (Davinci-002) & 56.8 & 50.2 & 48.8 & 51.6 & 52.5 & 51.9 & 51.2 & 41.9 & 51.2 & 51.6 & 50.5 & 49.3           \\ 
GPT-3 (Babbage-002) & 50.0$\dag$ & 50.0$\dag$ & 50.0$\dag$ & 50.0$\dag$ & 47.5 & 49.6 & 3.6* & 7.6* & 1.2* & 19.6* & 18.5* & 9.8*\\ 
GPT-3.5 & 49.6 & 51.2 & 50.4 & 50.0 & 52.0 & 50.6 & 43.6 & 50.4 & 53.6 & 50.0 & 47.5 & 49.1\\ 
GPT-4o mini & 50.0$\dag$ & 50.0$\dag$ & 50.0$\dag$ & 50.0$\dag$ & 52.5 & 50.4 & 57.2 & 66.4 & 60.0 & 63.2 & 50.0 & 59.8 \\ 
GPT-4o & 50.4 & 54.4 & 50.4 & 54.8 & 54.0 & 52.8 & 80.4 & \textbf{72.4} & \textbf{80.8} & \textbf{81.6} & 60.5 & \textbf{75.8} \\
Claude-3 (Sonnet) & 50.4 & 48.8 & 50.0 & 50.8 & \textbf{59.5} & 51.6 & 59.2 & 52.0 & 64.4 & 60.0 & \textbf{65.5} & 59.0\\ 
Claude-3.5 (Haiku) & 28.4 & 24.0 & 43.6 & 54.0 & 51.0 & 39.8 & 60.4 & 65.6 & 67.2 & 66.0 & 61.0 & 64.2\\ 
Gemini-1.5-flash & \textbf{75.2} & \textbf{65.6} & \textbf{67.2} & \textbf{76.0} & 53.0 & \textbf{68.0} & 72.4 & 70.8 & 72.4 & 78.4 & 58.5 & 71.0\\ 
Gemini-1.5-flash-8b & 50.0$\dag$ & 50.0$\dag$ & 50.0$\dag$ & 50.0$\dag$ & 52.5 & 50.4 & 66.8 & 67.2 & 65.2 & 65.2 & 58.5 & 64.8 \\ 
Deepseek-V3 & 50.4 & 50.4 & 50.0 & 50.0 & 60.5 & 51.9 & \textbf{80.8} & 70.4 & 76.4 & 77.6 & 63.5 & 74.2 \\
\hline
\end{tabular}
\caption{Model accuracy of standard method and CausalCoT across different reasoning categories. Note: * The average accuracy is only $9.8\%$ because most of responses are not ``Yes'' or ``No'' but ``incomprehensible'', which means LLM cannot follow instruction of CausalCoT instruction well to infer. More details will be explained in the Appendix H. $\dag$ indicates that the LLM predicts all questions as either ``Yes'' or ``No'', leading to a 50\% accuracy. }
\label{table:model_comparison}
\end{table*}

\paragraph{Basic Counterfactual.}  
The basic counterfactual type addresses simple ``what-if'' scenarios. In this scenario, it is formalized as $Y_x(u)$, which serves as a potential outcome expression. In causal reasoning, potential outcomes refer to the hypothetical results observed when a variable is set to a particular value \cite{holland1986statistics}. In $Y_x(u)$ expression, $Y$ is the outcome variable,$x$ is the value considered in the hypothetical scenario, and $u$ denotes the context. To illustrate, consider a lawn irrigation system that only activates when the weather is sunny and the soil is dry. In this example, $Y_x(u)$ describes the system’s potential behavior when the weather condition x is imposed while the soil condition (context $u$) remains unchanged. Consequently, when asking whether the system would activate if the weather changed to cloudy, the relevant counterfactual outcome is $Y_{cloudy}(u)$.

\paragraph{Joint Counterfactual.}  
This type involves a counterfactual scenario in which multiple variables are set simultaneously. Formally, it is expressed as $Y_{x,z}(u)$, representing the outcome $Y$ after setting $X = x$ and $Z = z$. For instance, a lawn irrigation system will activate if the weather is sunny, but it also requires dry soil as a trigger. Suppose $Z$ represents the weather condition and $X$ represents the soil moisture condition. The query asks if the irrigation system will activate when the weather changes to cloudy and the sensor detects that the soil is moist meanwhile. The relevant counterfactual outcome is $Y_{cloudy, moist}(u)$. This scenario examines the combined effect of both actions happening simultaneously.

\paragraph{Nested Counterfactual.}  
Nested counterfactual involves sequential dependencies between variables. This is represented as $Y_{Z_x}(u)$, where an intervention on $X$ affects $Z$, which in turn impacts $Y$. For example, if the weather had been cloudy, which is a counterfactual weather state, and under this scenario, the sensor detected moist soil instead of dry soil, would the irrigation system activate? In this framework, $Z$ represents the weather condition, $X$ is the soil moisture reading. $Z$ depends on the counterfactual value of $X$ through the system’s structural causal relationships.

\paragraph{Conditional Counterfactual.}  
This type introduces observed conditions into the counterfactual world \cite{pearl2009causality}. Formally, it is written as $Y_x(u) \mid Z_x(u) = z$, asking how $Y$ would change if $X$ was set to $x$ while $Z = z$ being observed as a condition. For example, a lawn irrigation system will activate if the weather is sunny, but it also requires dry soil as a trigger. Now the weather is observed as sunny, the query evaluates whether the lawn irrigation system still activate or not if the sensor detects moist soil instead of dryness. Here, $Z = z$ represents the weather is observed as sunny, which is a given condition for reasoning. 


\paragraph{Backdoor Counterfactual.}
This type involves counterfactual reasoning in the presence of backdoor paths that create confounding between the treatment variable and the outcome \cite{pearl2009causality}. Formally, it addresses queries of the form $Y_x(u)$ when there exist backdoor paths from $X$ to $Y$ through confounders. In such scenarios, the causal effect cannot be directly identified without controlling for the confounding variables along the backdoor paths. For example, a manager considers evaluating whether a new marketing campaign would increase sales if it is implemented. However, both the decision to launch the campaign and the sales outcome might be influenced by seasonal demand patterns. Here, the backdoor path runs from the marketing campaign through seasonal demand to sales, creating a spurious association. The counterfactual query ``Would sales increase if we launched the campaign?'' requires accounting for this confounding by either controlling for seasonal effects or using other identification strategies. In our dataset, backdoor counterfactuals test whether LLMs can distinguish between genuine causal effects and spurious correlations when reasoning about alternative scenarios.

\subsection{Query Generation and Quality Assessment}

Each query consists of background information and a specific question. Table~\ref{tab:counterfactual_types} illustrates how samples are generated using various deterministic counterfactual query types. The background is constructed with causal graphs and story templates, and variable names are replaced by nonsensical, artificially generated words (e.g., “Kelp,” “Ziklo”) to prevent models from relying on memorized knowledge. In this way, we force LLMs to engage in causal reasoning rather than using prior knowledge in pretraining data. The dataset also features balanced distributions in multiple dimensions, with binary responses evenly split between $50\%$ “Yes” and $50\%$ “No.” This balance extends across different question types and difficulty levels, ensuring a uniform response distribution within each category. The dataset consists of 1,200 questions, categorized into five distinct types, with each type containing 200 or 250 questions. Within each type, there is an equal distribution of answers, comprising 100 ``Yes'' responses and 100 ``No'' responses or 125 ``Yes'' and 125 ``No''. Additionally, the dataset is stratified based on five levels of difficulty, determined by the number of events present in each question, ranging from 5 to 9. Each difficulty level includes 240 questions, maintaining a balanced distribution of answers with 120 ``Yes'' and 120 ``No''. We also conduct a human evaluation on these queries, with further details provided in Appendix B.

\section{Benchmarking LLMs on CounterBench}

We conducted comprehensive experiments to systematically evaluate the performance of current LLMs on counterfactual reasoning tasks, demonstrating their capabilities using state-of-the-art reasoning techniques.

\noindent \textbf{Models.}
The tested LLM models include GPT-3.5 turbo, GPT-4o, GPT-4o mini, Davinci-002, Babbage-002 \cite{OpenAI2024models}, Claude 3.5 Haiku, Claude 3 Sonnet \cite{anthropic2024api}, Deepseek-V3 \cite{DeepSeek2024} and Gemini-1.5-Flash and Gemini-1.5-Flash-8B \cite{google2024gemini}.

\noindent \textbf{Reasoning Strategies.}
In our baseline evaluations, we employed two distinct reasoning strategies to assess these models. The first relied on standard prompting methods without specialized instructions. The second used the advanced CausalCoT approach \cite{jin2023cladder}, an extension of the Chain-of-Thought prompting paradigm \cite{wei2022chain}. By integrating a systematic derivation process, including causal graph extraction, query type classification, data collection, and formalization, CausalCoT ensures robust logical consistency and high reasoning accuracy.

\noindent \textbf{Evaluation Settings.}
Within our evaluation framework, responses are classified into three distinct categories, ``Yes'', ``No'', and ``Incomprehensible''. The latter encompassing responses that are either ambiguous or lack clear meaning, typically manifesting when no answer is detected, such as NULL returns or mere query echoes. During inference, we set the temperature at 0. We employ inference accuracy as our primary performance metric.

\subsection{Experimental Results}
As shown in Table~\ref{table:model_comparison}, the results indicate that without specific instructions, most LLMs struggle with counterfactual reasoning, performing no better than random guessing in terms of accuracy. Specifically, for model GPT-4o mini, we observed consistent predictions of either ``Yes'' or ``No,'' resulting in a $50.0\%$ accuracy in the first four kinds of questions. Among all tested models, Gemini-1.5-flash achieved the highest baseline performance with an accuracy of $68.0\%$. Although the CausalCoT approach is designed to enhance the causal reasoning capabilities of LLMs, our empirical findings suggest that it does not significantly improve their performance in counterfactual reasoning tasks. Most models showed minimal or no improvement, as exemplified by GPT-3.5 Turbo. The best performance model in the CausalCoT is GPT-4o, achieved an accuracy of only $75.8\%$.

\begin{figure}[htbp]
  \centering
  \includegraphics[width=0.4\textwidth]{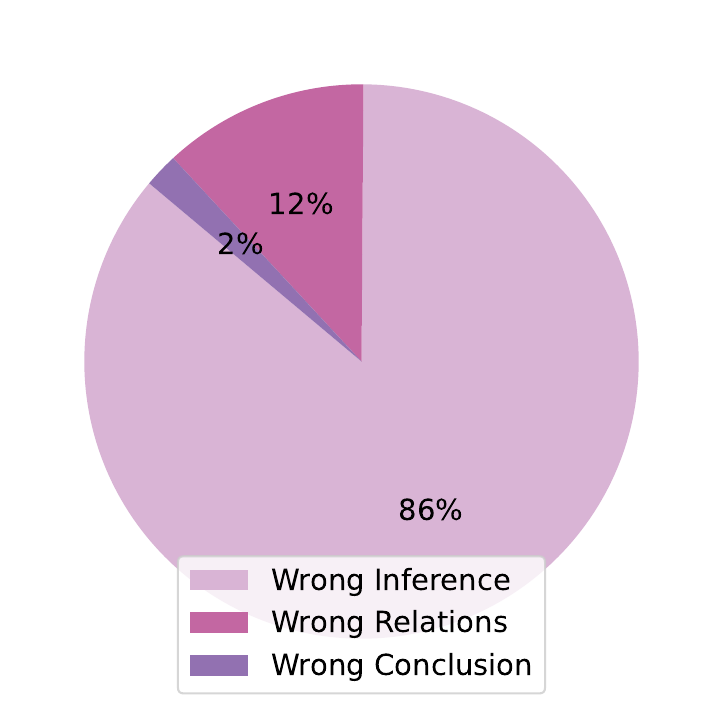}
  \caption{Error Analysis of CausalCoT.}
  \label{fig:pie}
\end{figure}

\noindent \textbf{Error Analysis.}
To systematically analyze the limitations of existing approaches, we conducted an error analysis on responses generated by CausalCoT. Our analysis focuses on three key components: causal data collection, inference process, and conclusion derivation. Through careful examination of each component, we identified three primary categories of errors:  \textbf{Wrong causal relationships:} This error occurs when LLMs cannot construct accurate causal graphs or extract known values from background information.  
\textbf{Wrong inference process:} This happens when LLMs, despite correctly identifying causal relationships, make incorrect predictions of the target event $Y$. \textbf{Wrong conclusion:} This type of error arises when LLMs reach contradictory final answers, even after correctly computing the value of $Y$. The distribution of these error categories is illustrated in Figure ~\ref{fig:pie}. Notably, $86\%$ of errors occur in the inference process, revealing that even with well-constructed causal graphs, LLMs struggle significantly with deriving accurate predictions through reasoning. 


\section{Proposed Reasoning Strategy}

As discussed in previous section, the primary challenge for large language models is to minimize incorrect inferences, which are a major source of errors. To address this challenge, we propose \textbf{CoIn} (Counterfactual Inference), a systematic reasoning framework that guides large language models through structured problem-solving instead of relying on intuitive shortcuts or memorized patterns. Our approach transforms counterfactual queries into a five-phase algorithmic process, mirroring how humans naturally approach ``what if'' questions \cite{sel2023algorithm}: first understanding what actually happened, then imagining the alternative scenario, systematically working through the consequences, and finally double-checking the logic. This structured approach significantly reduces reasoning errors by breaking down complex problems into manageable steps with built-in validation. An example of proposed paradigm is provided in Appendix A.

The CoIn framework consists of five key phases:
\textbf{Extraction: } Extract Counterfactual Information from the given natural language facts. 
\textbf{Abduction:} Infer the underlying conditions from observed facts;
\textbf{Intervention Action:} Apply the hypothetical changes specified in the query;
\textbf{Forward Inference:} Systematically trace through the causal consequences;
\textbf{Back-tracking Validation:} Verify the logical consistency of the entire reasoning chain. Each phase serves a specific purpose in ensuring accurate counterfactual reasoning, and they provide a robust methodology for handling complex causal dependencies together. Below, we describe each phase in detail, explaining its role and how it contributes to the overall process.

\subsection{Extraction} In the first phase, we focus on systematically gathering all relevant information explicitly stated in the scenario. The process begins with constructing the causal graph by identifying relationships between events and representing them in a clear ``$event\ 1 \to event\ 2$'' format, which eliminates potential ambiguities. Next, we collect the given values for each variable from both background information and questions, where these values indicate whether specific events occur or not. Crucially, this phase maintains strict adherence to explicitly stated information, avoiding any unsupported inferences or assumptions in favor of a rigorous and unbiased data collection process. 
\subsection{Abduction}
This phase focuses on inferring the posterior constraints over the exogenous noise variables, equivalently, constraints over parent assignments that make the observed factual world consistent with the structural equations. For each observed variable $V$ with value $v_{\mathrm{obs}}$, we invert its structural equation
$V := f_V(\mathrm{Parents}(V), U_V)$
to obtain either a unique solution for $U_V$ or a feasible set over $U_V$ given the parents. In deterministic logical models, this is often conveniently carried out by deducing parent assignments that must hold for $v_{\mathrm{obs}}$ to be true. The resulting values are stored as the factual world knowledge base and will be held fixed in subsequent Intervention Action and Forward Inference, ensuring counterfactuals are evaluated in the same world.


\subsection{Intervention Action}
In this phase, the framework applies the counterfactual interventions described in the query. This involves modifying the original set of causal rules by replacing the equations for the intervened variables with constant values, resulting in an updated set of rules. The interventions are incorporated into the knowledge base, effectively making a precise alteration to the causal graph. This phase captures the core ``what if'' element of the query, allowing the framework to simulate hypothetical worlds in a controlled way. It focuses on specific changes, which streamlines the exploration by limiting the search to paths directly affected by the intervention, similar to how efficient searches eliminate unnecessary branches.

\subsection{Forward Inference}

During this iterative phase, the framework predicts values for unobserved variables by selecting nodes in the causal graph whose parent variables are already known in the knowledge base and evaluating their updated equations. Beginning with the intervened variables and the inferred noise terms, it gradually computes the effects on downstream variables until it reaches the target variable $Y$. If a node's value cannot be calculated due to missing information about its parents, the framework chooses another suitable node and continues the process until $Y$ is determined. This forward progression mimics a depth-first exploration of causal chains, enabling the framework to dynamically construct and assess potential outcomes. By focusing on nodes that can be computed immediately, it navigates the dependency graph efficiently, steering clear of unproductive paths and promoting a methodical advancement toward the solution. The details of this phase is in the Appendix G.

\subsection{Back-tracking Validation}

To confirm that the predicted values are logically consistent, this final phase retraces the steps through the knowledge base and re-evaluates the equation for each non-noise variable using the predicted values. For every such variable $V$, it recalculates the expected value based on the updated equation and checks if it matches the previously stored value. If any mismatch occurs, the framework signals an error, highlighting a potential issue in the earlier reasoning that may need reevaluation. This validation serves as a protective measure against errors that could accumulate during the process, akin to retracing a path to confirm its validity. 

\section{Experiments}
\subsection{Experiment Setup}
We adopted the same LLMs as mentioned in Section 3 for our experiments. To establish baselines, we implemented both CausalCoT \cite{jin2023cladder} and standard solver strategies. The latter integrates external tools into the chain-of-thought process, as described in \cite{hua2024improving}. Specifically, this approach combines LLMs with Structural Causal Model (SCM) tools \cite{pearl2009causality} for causal inference. The study introduces CausalTool, a suite of 10 inference tools designed for various causal tasks. It leverages LLMs to classify causal questions, extract causal graphs and formalized data, and route them to the appropriate tools for inference, with the final answer generated by the LLM. \footnote{Since the source code for CausalTool is not publicly available, we re-implemented its counterfactual inference procedure based on descriptions in the original paper.} During inference, the temperature is set to zero.
\subsection{Main Result}

\begin{table}[t]
\centering
\setlength{\tabcolsep}{0.1mm}
\scalebox{0.75}{
\setlength{\tabcolsep}{1pt} 
\begin{tabular}{p{3.2cm}>{\centering\arraybackslash}p{1.8cm}>{\centering\arraybackslash}p{1.8cm}>{\centering\arraybackslash}p{1.2cm}>{\centering\arraybackslash}p{0.9cm}}
\toprule
\rowcolor{mygray}
\textbf{Model} & \textbf{Standard} & \textbf{CausalCoT} & \textbf{Solver} & \textbf{Ours}\\
\midrule
GPT-3 (Davinci-002) & \textbf{51.9} & 49.3 & 50.1 & 49.6\\
GPT-3 (Babbage-002) & \textbf{49.6} & 9.8 & 47.9 & 45.8\\
GPT-4o mini & 50.4 & 59.8 & 47.2 & \textbf{79.9}\\
GPT-4o & 52.8 & 75.8 & 51.4 & \textbf{89.4}\\
GPT-3.5 turbo & 50.6 & 49.1 & 49.6 & \textbf{58.9}\\
Claude-3 (Sonnet) & 51.6 & 59.0 & 51.8 & \textbf{89.8}\\
Claude-3.5 (Haiku) & 39.8 & 64.2 & 48.3 & \textbf{79.1}\\
Gemini-1.5-flash & 68.0 & 71.0 & 52.8 & \textbf{89.9}\\
Gemini-1.5-flash-8b & 50.4 & 64.8 & 50.3 & \textbf{83.9}\\
Deepseek-V3 & 51.9 & 74.2 & 49.3 & \textbf{91.8}\\
\bottomrule
\end{tabular}
}
\caption{Model accuracy on CounterBench. We report the average accuracy for four inference strategies: Standard, CausalCoT, Solver, and CoIn.}
\label{tab:model_accuracy}
\end{table}

The comprehensive performance comparison across all models is presented in Table~\ref{tab:model_accuracy}. Our approach demonstrates notable improvements over existing methods across the model spectrum, with particularly noteworthy performance gains achieved by smaller language models, including GPT-4o mini, Claude-3.5 Haiku, and Gemini-1.5-flash-8b. For instance, our method enables GPT-4o mini to achieve an accuracy of $79.9\%$, surpassing the performance of several larger models without CoIn enhancement. As detailed in Table~\ref{tab:model_accuracy_type}, taking GPT-4o mini as an example, CoIn achieves superior performance across all five types of counterfactual questions, with particularly better results on basic questions compared to more complex variants. Additionally, state-of-the-art LLMs such as GPT-4o, Gemini-1.5-flash, and Deepseek-V3 achieve remarkable accuracy approaching or exceeding $90\%$ when augmented with our method. Taken GPT-4o as example, Our strategy improves the accuracy of the model from $75.8\%$ to $89.4\%$, demonstrating CoIn's effectiveness in guiding LLMs through algorithm to explore paths step-by-step. The results indicate that contemporary LLMs, when equipped with our strategy, can effectively resolve most formal complex counterfactual problems. The details of all performance of our results are presented in Appendix C. Moreover, in Appendix D, we conducted error analysis. The analysis reveals a substantial reduction in errors of inference process. Moreover, we also examine the impact of complex causal relationships on outcomes. We found that accuracy decreases as the number of variants increases for CausalCoT and CoIn. The details are shown in the next section. 
\begin{table}[t]
\centering
\scalebox{0.9}{
\setlength{\tabcolsep}{1pt} 
\begin{tabular}{p{1.8cm}>{\centering\arraybackslash}p{1cm}>{\centering\arraybackslash}p{1cm}>{\centering\arraybackslash}p{1cm}>{\centering\arraybackslash}p{1.2cm}>{\centering\arraybackslash}p{1cm}>{\centering\arraybackslash}p{1cm}}
\toprule
\rowcolor{mygray}
\textbf{Methods} & \textbf{Basic} & \textbf{Cond.} & \textbf{Joint} & \textbf{Nested} & \textbf{Back.} & \textbf{Avg.} \\
\midrule
Standard & 50.0 & 50.0 & 50.0 & 50.0 & 52.5 & 50.4\\
CausalCoT & 57.2 & 66.4 & 60.0 & 63.2 & 50.0 & 59.8 \\
Solver & 35.2 & 54.4 & 50.4 & 50.0 & 45.5 & 47.2 \\
Ours & \textbf{82.8} & \textbf{79.2} & \textbf{80.0} & \textbf{80.4} & \textbf{76.5} & \textbf{79.9} \\
\bottomrule
\end{tabular}
}
\caption{Accuracy of GPT-4o mini across five query types in the CounterBench.}
\label{tab:model_accuracy_type}
\end{table}

\subsection{Validating Generalization Ability}
In this section, we evaluate the generalization capability of CoIn using the CLADDER dataset \cite{jin2023cladder}. CLADDER is a dataset focus on the causal reasoning questions. We utilize the CLADDER dataset to determine if the proposed method can be extended beyond the CounterBench dataset. Unlike our dataset, which focuses on formal rules, CLADDER includes examples that utilize common sense knowledge rather than causal inference abilities to answer queries. Specifically, it encompasses both commonsense and anti-commonsense scenarios, allowing us to explore whether CoIn remains effective under the influence of pretraining knowledge in LLMs. We conduct experiments on the counterfactual subset of CLADDER. We applied the Gemini-1.5 and Gemini-1.5-8b models to both commonsense and anti-commonsense queries, with the results displayed in Figure~\ref{fig:structure1} and detailed examples in Appendix E. These results demonstrate that performance is consistently stable across different reasoning paradigms, suggesting that pretraining knowledge has a limited impact on the CoIn counterfactual reasoning capabilities. Furthermore, our method achieves an accuracy of 78.98\%, outperforming both CausalCoT at 64.77\% and the Standard method at 64.20\%. This performance underscores CoIn's generalizability and its potential for broader application in various counterfactual reasoning tasks.

\begin{figure}[htbp]
  \centering
  
  \includegraphics[width=0.47\textwidth]{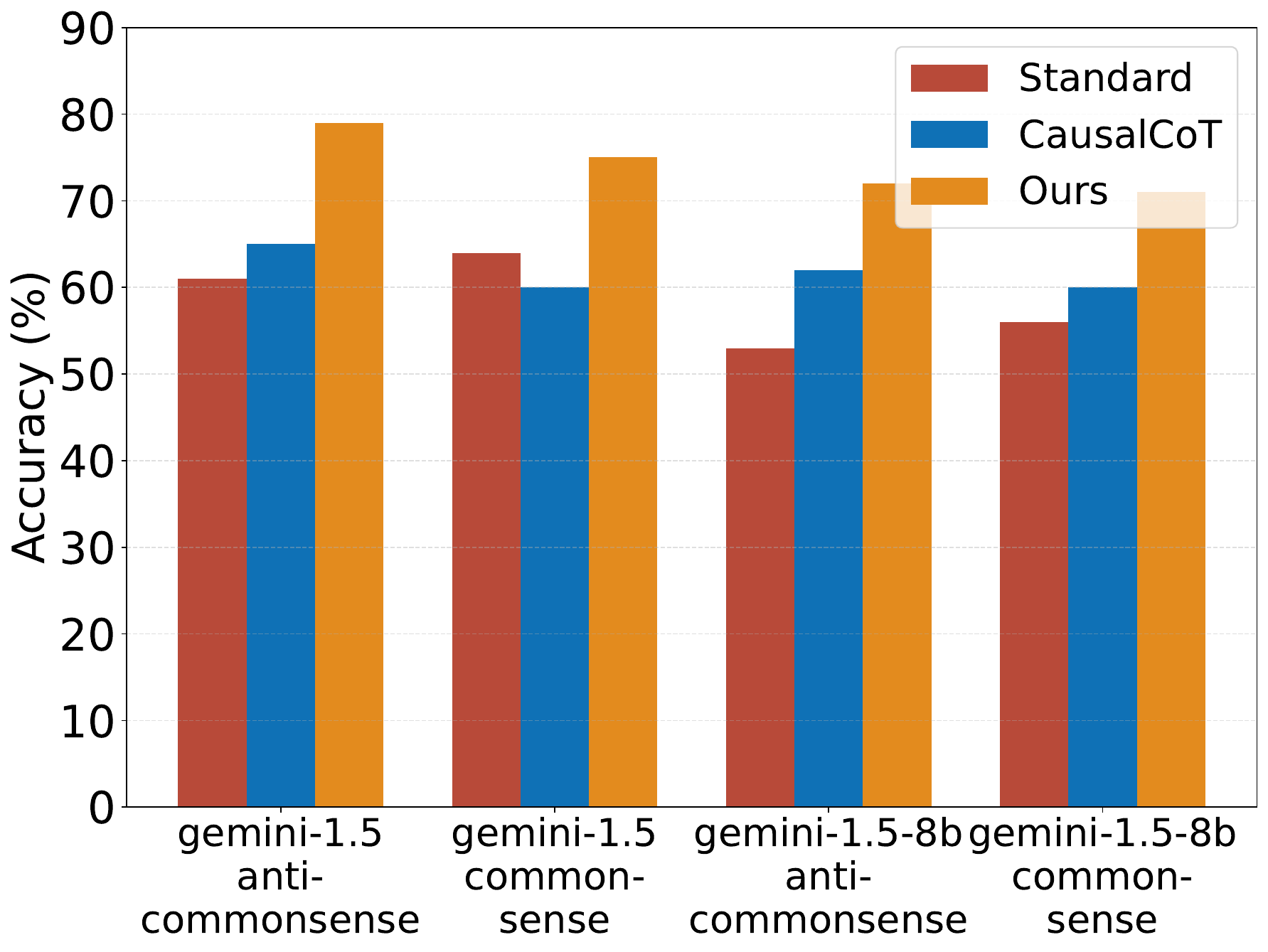}
  \caption{Accuracy comparison between Standard, CoIn, and CausalCoT method in Anti-commonsense and Commonsense Dataset.}
  \label{fig:structure1}
\end{figure}

\section{Related Work}

\paragraph{Counterfactual Reasoning.}
Counterfactual reasoning explores how outcomes change when certain variables are altered from their historical states. In Structural Causal Models (SCMs), Pearl’s \cite{pearl2009causality} “surgery” and do-calculus provide systematic ways to infer intervention outcomes, highlighting deep causal knowledge required for accurate inference. Counterfactuals can be deterministic or probabilistic: deterministic settings yield predictable outcomes from given conditions, while probabilistic models incorporate inherent uncertainties. These methods have gained traction in domains like social sciences, where they assess alternative policy outcomes and study causal mechanisms in observational data \cite{morgan2015counterfactuals}, and in medicine, where they enable personalized treatment and decision support \cite{johansson2016learning,shalit2017estimating,louizos2017causal,yoon2018ganite}. In artificial intelligence, counterfactual reasoning is crucial for interpretability and fairness, enabling models to generate alternative scenarios and assess decision-making robustness. Although recent efforts extend counterfactual reasoning to LLMs \cite{jin2023cladder}, significant challenges persist, particularly regarding complex variable relationships in high-dimensional text data. Consequently, bridging the gap between textual complexity and robust causal inference remains a focal point for future research.

\paragraph{LLMs in Counterfactual Learning.}
With the rapid evolution of LLMs, the research community has increasingly focused on their ability to perform causal inference \cite{zhang2023understanding,ashwani2024cause}. A prominent example is Causal Agent, an agent-based LLM framework that merges an LLM with causal tools for complex tasks \cite{han2024causal}. While it excels at identifying causal associations and conducting interventions, it largely omits counterfactual reasoning, limiting its applicability to more advanced scenarios. Current efforts to integrate counterfactual reasoning into LLMs typically follow two paths. First, commonsense-based approaches leverage background knowledge to imagine scenarios that defy established facts \cite{ning2024skeleton,chatzi2024counterfactual,musi2024fallacy,vicuna2023}, such as positing alternative historical outcomes. Second, graph-based methods employ formal causal graphs and external Python packages for computations, as seen in CausalTool \cite{hua2024improving}. Although these methods effectively incorporate structured causal information, they often offload key calculations outside the LLM. 

\section{Conclusion}
In this work, we develop and extend CounterBench, a counterfactual reasoning dataset with five problem types for LLM evaluation. Our findings reveal that most LLMs perform near-randomly, with state-of-the-art methods showing minimal improvement. To address these challenges, we propose CoIn, a reasoning paradigm inspired by formal causal inference principles and planning strategies. CoIn guides LLMs through iterative thinking and backtracking to explore reasoning paths more effectively. Our approach significantly enhances counterfactual reasoning capabilities of LLMs.

\bibliography{reference}

@article{sel2023algorithm,
  title={Algorithm of thoughts: Enhancing exploration of ideas in large language models},
  author={Sel, Bilgehan and Al-Tawaha, Ahmad and Khattar, Vanshaj and Jia, Ruoxi and Jin, Ming},
  journal={arXiv preprint arXiv:2308.10379},
  year={2023}
}

@book{pearl2018book,
  title={The book of why: the new science of cause and effect},
  author={Pearl, Judea and Mackenzie, Dana},
  year={2018},
  publisher={Basic books}
}

@inproceedings{jin2023cladder,
  title={Cladder: Assessing causal reasoning in language models},
  author={Jin, Zhijing and Chen, Yuen and Leeb, Felix and Gresele, Luigi and Kamal, Ojasv and Zhiheng, LYU and Blin, Kevin and Adauto, Fernando Gonzalez and Kleiman-Weiner, Max and Sachan, Mrinmaya and others},
  booktitle={Thirty-seventh conference on neural information processing systems},
  year={2023}
}

@article{ma2024causal,
  title={Causal inference with large language model: A survey},
  author={Ma, Jing},
  journal={arXiv preprint arXiv:2409.09822},
  year={2024}
}

@article{kiciman2023causal,
  title={Causal reasoning and large language models: Opening a new frontier for causality},
  author={K{\i}c{\i}man, Emre and Ness, Robert and Sharma, Amit and Tan, Chenhao},
  journal={arXiv preprint arXiv:2305.00050},
  year={2023}
}

@article{zevcevic2023causal,
  title={Causal parrots: Large language models may talk causality but are not causal},
  author={Ze{\v{c}}evi{\'c}, Matej and Willig, Moritz and Dhami, Devendra Singh and Kersting, Kristian},
  journal={arXiv preprint arXiv:2308.13067},
  year={2023}
}

@article{liu2024large,
  title={Large language models and causal inference in collaboration: A comprehensive survey},
  author={Liu, Xiaoyu and Xu, Paiheng and Wu, Junda and Yuan, Jiaxin and Yang, Yifan and Zhou, Yuhang and Liu, Fuxiao and Guan, Tianrui and Wang, Haoliang and Yu, Tong and others},
  journal={arXiv preprint arXiv:2403.09606},
  year={2024}
}

@article{wei2022chain,
  title={Chain-of-thought prompting elicits reasoning in large language models},
  author={Wei, Jason and Wang, Xuezhi and Schuurmans, Dale and Bosma, Maarten and Xia, Fei and Chi, Ed and Le, Quoc V and Zhou, Denny and others},
  journal={Advances in neural information processing systems},
  volume={35},
  pages={24824--24837},
  year={2022}
}

@article{zhang2023understanding,
  title={Understanding causality with large language models: Feasibility and opportunities},
  author={Zhang, Cheng and Bauer, Stefan and Bennett, Paul and Gao, Jiangfeng and Gong, Wenbo and Hilmkil, Agrin and Jennings, Joel and Ma, Chao and Minka, Tom and Pawlowski, Nick and others},
  journal={arXiv preprint arXiv:2304.05524},
  year={2023}
}

@inproceedings{ashwani2024cause,
  title={Cause and effect: Can large language models truly understand causality?},
  author={Ashwani, Swagata and Hegde, Kshiteesh and Mannuru, Nishith Reddy and Sengar, Dushyant Singh and Jindal, Mayank and Kathala, Krishna Chaitanya Rao and Banga, Dishant and Jain, Vinija and Chadha, Aman},
  booktitle={Proceedings of the AAAI Symposium Series},
  volume={4},
  number={1},
  pages={2--9},
  year={2024}
}

@article{chatzi2024counterfactual,
  title={Counterfactual token generation in large language models},
  author={Chatzi, Ivi and Benz, Nina Corvelo and Straitouri, Eleni and Tsirtsis, Stratis and Gomez-Rodriguez, Manuel},
  journal={arXiv preprint arXiv:2409.17027},
  year={2024}
}

@inproceedings{ning2024skeleton,
  title={Skeleton-of-thought: Prompting LLMs for efficient parallel generation},
  author={Ning, Xuefei and Lin, Zinan and Zhou, Zixuan and Wang, Zifu and Yang, Huazhong and Wang, Yu},
  booktitle={The Twelfth International Conference on Learning Representations},
  year={2024}
}

@inproceedings{musi2024fallacy,
  title={The Fallacy of Explainable Generative AI: evidence from argumentative prompting in two domains},
  author={Musi, Elena and Palmieri, Rudi},
  booktitle={CEUR Workshop Proceedings},
  volume={3769},
  pages={59--69},
  year={2024}
}

@misc{vicuna2023,
  title = {Vicuna: An open-source chatbot impressing {GPT-4} with 90\%* {ChatGPT} quality},
  author = {{Vicuna}},
  year = {2023},
  howpublished = {\url{https://vicuna.lmsys.org/}},
  note = {Accessed: 2023}
}

@book{pearl2009causality,
  title={Causality},
  author={Pearl, Judea},
  year={2009},
  publisher={Cambridge university press}
}

@book{morgan2015counterfactuals,
  title={Counterfactuals and causal inference},
  author={Morgan, SL},
  year={2015},
  publisher={Cambridge University Press}
}

@inproceedings{johansson2016learning,
  title={Learning representations for counterfactual inference},
  author={Johansson, Fredrik and Shalit, Uri and Sontag, David},
  booktitle={International conference on machine learning},
  pages={3020--3029},
  year={2016},
  organization={PMLR}
}

@inproceedings{shalit2017estimating,
  title={Estimating individual treatment effect: generalization bounds and algorithms},
  author={Shalit, Uri and Johansson, Fredrik D and Sontag, David},
  booktitle={International conference on machine learning},
  pages={3076--3085},
  year={2017},
  organization={PMLR}
}

@article{louizos2017causal,
  title={Causal effect inference with deep latent-variable models},
  author={Louizos, Christos and Shalit, Uri and Mooij, Joris M and Sontag, David and Zemel, Richard and Welling, Max},
  journal={Advances in neural information processing systems},
  volume={30},
  year={2017}
}

@inproceedings{yoon2018ganite,
  title={GANITE: Estimation of individualized treatment effects using generative adversarial nets},
  author={Yoon, Jinsung and Jordon, James and Van Der Schaar, Mihaela},
  booktitle={International conference on learning representations},
  year={2018}
}

@misc{Anthropic2024api,
author = {Anthropic},
title = {Claude},
howpublished = {\url{https://www.anthropic.com/api}},
year = {2024},
note = {Accessed: 2025-01-06}
}

@misc{Google2024gemini,
author = {Google},
title = {Gemini},
year = {2024},
howpublished = {\url{https://gemini.google.com/}},
note = {Accessed: 2025-01-06}
}

@misc{OpenAI2024models,
author = {OpenAI},
title = {Models},
year = {2024},
howpublished = {\url{https://platform.openai.com/docs/models}},
note = {Accessed: 2025-01-06}
}

@inproceedings{liu2024proficient,
  title={How Proficient Are Large Language Models in Formal Languages? An In-Depth Insight for Knowledge Base Question Answering},
  author={Liu, Jinxin and Cao, Shulin and Shi, Jiaxin and Zhang, Tingjian and Nie, Lunyiu and Hu, Linmei and Hou, Lei and Li, Juanzi},
  booktitle={Findings of the Association for Computational Linguistics ACL 2024},
  pages={792--815},
  year={2024}
}

@article{petroni2019language,
  title={Language models as knowledge bases?},
  author={Petroni, Fabio and Rockt{\"a}schel, Tim and Lewis, Patrick and Bakhtin, Anton and Wu, Yuxiang and Miller, Alexander H and Riedel, Sebastian},
  journal={arXiv preprint arXiv:1909.01066},
  year={2019}
}

@article{liang2024mapping,
  title={Mapping the increasing use of llms in scientific papers},
  author={Liang, Weixin and Zhang, Yaohui and Wu, Zhengxuan and Lepp, Haley and Ji, Wenlong and Zhao, Xuandong and Cao, Hancheng and Liu, Sheng and He, Siyu and Huang, Zhi and others},
  journal={arXiv preprint arXiv:2404.01268},
  year={2024}
}

@article{tarassow2023potential,
  title={The potential of LLMs for coding with low-resource and domain-specific programming languages},
  author={Tarassow, Artur},
  journal={arXiv preprint arXiv:2307.13018},
  year={2023}
}

@misc{li2024relationfirstmodelingparadigmcausal,
      title={Relation-First Modeling Paradigm for Causal Representation Learning toward the Development of AGI}, 
      author={Jia Li and Xiang Li},
      year={2024},
      eprint={2307.16387},
      archivePrefix={arXiv},
      primaryClass={cs.AI},
      url={https://arxiv.org/abs/2307.16387}, 
}

@misc{sahota2023causalai,
  title = {Causal AI: Bridging the Gap Between Correlation and Causation},
  author = {Sahota, Neil},
  howpublished = {\href{https://www.neilsahota.com/causal-ai-bridging-the-gap-between-correlation-and-causation/}{https://www.neilsahota.com}},
  note = {Accessed: 2025-01-06},
  year = {2023}
}

@misc{alwin2023causalai,
  title = {Understanding Causal AI: Bridging The Gap Between Correlation And Causation},
  author = {Alwin},
  howpublished = {\url{https://www.alwin.io/causal-ai}},
  note = {Accessed: 2025-01-06},
  year = {2023}
}

@article{gvozdenovic2021causal,
  title={Causal inference concepts applied to three observational studies in the context of vaccine development: from theory to practice},
  author={Gvozdenovi{\'c}, Emilia and Malvisi, Lucio and Cinconze, Elisa and Vansteelandt, Stijn and Nakanwagi, Phoebe and Aris, Emmanuel and Rosillon, Dominique},
  journal={BMC Medical Research Methodology},
  volume={21},
  pages={1--10},
  year={2021},
  publisher={Springer}
}

@article{kyrimi2025counterfactual,
  title={Counterfactual reasoning using causal Bayesian networks as a healthcare governance tool},
  author={Kyrimi, Evangelia and Mossadegh, Somayyeh and Wohlgemut, Jared M and Stoner, Rebecca S and Tai, Nigel RM and Marsh, William},
  journal={International Journal of Medical Informatics},
  volume={193},
  pages={105681},
  year={2025},
  publisher={Elsevier}
}

@inproceedings{kasirzadeh2021use,
  title={The use and misuse of counterfactuals in ethical machine learning},
  author={Kasirzadeh, Atoosa and Smart, Andrew},
  booktitle={Proceedings of the 2021 ACM Conference on Fairness, Accountability, and Transparency},
  pages={228--236},
  year={2021}
}

@article{krishnamurthy2002counterfactual,
  title={Counterfactual thinking and advertising responses},
  author={Krishnamurthy, Parthasarathy and Sivaraman, Anuradha},
  journal={Journal of Consumer Research},
  volume={28},
  number={4},
  pages={650--658},
  year={2002},
  publisher={The University of Chicago Press}
}

@article{loi2012note,
  title={A note on the impact evaluation of public policies: the counterfactual analysis},
  author={Loi, Massimo and Rodrigues, Margarida},
  year={2012}
}

@article{koonce2011judging,
  title={Judging the relevance of fair value for financial instruments},
  author={Koonce, Lisa and Nelson, Karen K and Shakespeare, Catherine M},
  journal={The Accounting Review},
  volume={86},
  number={6},
  pages={2075--2098},
  year={2011},
  publisher={American Accounting Assocation}
}

@article{gow2016causal,
  title={Causal inference in accounting research},
  author={Gow, Ian D and Larcker, David F and Reiss, Peter C},
  journal={Journal of Accounting Research},
  volume={54},
  number={2},
  pages={477--523},
  year={2016},
  publisher={Wiley Online Library}
}

@inproceedings{hua2024improving,
  title={Improving Causal Inference of Large Language Models with SCM Tools},
  author={Hua, Zhenyang and Xing, Shuyue and Jiang, Huixing and Wei, Chen and Wang, Xiaojie},
  booktitle={CCF International Conference on Natural Language Processing and Chinese Computing},
  pages={3--14},
  year={2024},
  organization={Springer}
}

@article{han2024causal,
  title={Causal agent based on large language model},
  author={Han, Kairong and Kuang, Kun and Zhao, Ziyu and Ye, Junjian and Wu, Fei},
  journal={arXiv preprint arXiv:2408.06849},
  year={2024}
}

@article{sharma2020dowhy,
  title={DoWhy: An end-to-end library for causal inference},
  author={Sharma, Amit and Kiciman, Emre},
  journal={arXiv preprint arXiv:2011.04216},
  year={2020}
}

@article{chen2020causalml,
  title={Causalml: Python package for causal machine learning},
  author={Chen, Huigang and Harinen, Totte and Lee, Jeong-Yoon and Yung, Mike and Zhao, Zhenyu},
  journal={arXiv preprint arXiv:2002.11631},
  year={2020}
}

@article{feder2022causal,
  title={Causal inference in natural language processing: Estimation, prediction, interpretation and beyond},
  author={Feder, Amir and Keith, Katherine A and Manzoor, Emaad and Pryzant, Reid and Sridhar, Dhanya and Wood-Doughty, Zach and Eisenstein, Jacob and Grimmer, Justin and Reichart, Roi and Roberts, Margaret E and others},
  journal={Transactions of the Association for Computational Linguistics},
  volume={10},
  pages={1138--1158},
  year={2022},
  publisher={MIT Press One Broadway, 12th Floor, Cambridge, Massachusetts 02142, USA~…}
}

@article{holland1986statistics,
  title={Statistics and causal inference},
  author={Holland, Paul W},
  journal={Journal of the American statistical Association},
  volume={81},
  number={396},
  pages={945--960},
  year={1986},
  publisher={Taylor \& Francis}
}

@misc{DeepSeek2024,
  author = "{DeepSeek}",
  title = "{DeepSeek: AI-Powered Search Engine}",
  year = {2024},
  url = {https://www.deepseek.com/},
  note = {Accessed: 2025-02-15}
}

@article{bubeck2023sparks,
  title={Sparks of artificial general intelligence: Early experiments with gpt-4},
  author={Bubeck, S{\'e}bastien and Chandrasekaran, Varun and Eldan, Ronen and Gehrke, Johannes and Horvitz, Eric and Kamar, Ece and Lee, Peter and Lee, Yin Tat and Li, Yuanzhi and Lundberg, Scott and others},
  journal={arXiv preprint arXiv:2303.12712},
  year={2023}
}

\setcounter{secnumdepth}{0} 
\appendix
\section{Appendix}
\subsection{A. Method Prompt Design}
Our prompt design, as illustrated in Figure~\ref{fig:prompt design}, has been carefully structured to optimize the interaction between our algorithm and Large Language Models (LLMs). At its core, the design incorporates multiple few-shot examples that serve as comprehensive demonstrations of our algorithm's operational framework. These examples are strategically selected to showcase various reasoning patterns and problem-solving approaches, providing LLMs with a robust foundation for understanding the algorithm's methodology. Within the prompt, we have meticulously detailed step-by-step instructions, creating a clear and structured information flow that guides the model through the reasoning process. This structured approach leverages LLMs' inherent in-context learning capabilities, enabling them to systematically explore solution paths rather than relying on intuitive responses. This approach not only enhances the model’s ability to break down problems and evaluate paths efficiently but also reducing the likelihood of generating erroneous intermediate steps and improving reasoning stability.
\label{sec:Method Prompt Design}
\begin{figure*}[h!]
  \centering
  \begin{mdframed}[backgroundcolor=white, linewidth=1pt, linecolor=black]
    \textbf{Question:} 

----------------\\
User: 
Imagine a self-contained, hypothetical world with only the following conditions, and without any unmentioned factors or causal relationships: Pexu has a direct effect on Kraz. Not Kraz and Pexu has a direct effect on hwax. Hwax has a direct effect on rukz. Kraz is unobserved. We know that pexu and not kraz causes hwax. pexu causes kraz. hwax causes rukz. We observed an individual is hwax. Would an individual is rukz if not kraz instead of kraz? Please follow the instructions step by step. 
Step 1. Extract the causal graph: Identify the causal graph that depicts the relationships in the scenario. Let X = kraz; V2 = pexu; V3 = hwax; Y = rukz.  The diagram should simply consist of edges denoted in "var1 $\to$ var2" format, separated by commas. If you get V1 $\to$ Not V2 for example, you need to replace into V1 $\to$ V2. 
Step 2. Abduction and Action for information collection: Collect all the directly given information into given values set. 1 means given observed in question or observed. 0 means given not in question. Do not assume or infer other variables values by relations. Then, describe relations about how multiple variables influence another variable; it can result in AND, OR, or NOT.
Step 3. Infer and Predict the Y by information step by step. 
Step 4. Based on the result from the Step3, derive the final answer. There is an identifiable answer.

Assistant: 

Step 1) Extract the causal graph: Identify the causal graph that depicts the relationships in the scenario. Let V2 = kraz; X = pexu; V3 = hwax; Y = rukz. The causal graph is V2$\to$V3, V2$\to$X, X$\to$V3,V3$\to$Y.\\
Step 2) Abduction and Action for information collection: All given values: V2 = 1, and X = 0 (by abduction, V2 occurs), X = 0 (by action, X not occur); Relations: Y: (NOT V2) AND V3, V3: X AND NOT V2.\\
Step 3) Trying a promising first operation:\\
	1. V3 = not X and V2 = 1\\
	- Y = V3 = 1 $\to$ found it !\\
	Backtracking the solution:\\
	Step 3a:\\
		V3 = not X and V2 = 1\\
	Step 3b:\\
		Y = V3 = 1\\
Step 4) Since the result for the Y is 1, the overall answer to the question is yes.\\
----------------

User: 
Imagine a self-contained, hypothetical world with only the following conditions, and without any unmentioned factors or causal relationships: Vank has a direct effect on Praf, Praf has a direct effect on Scud, Scud has a direct effect on Wrenk, Wrenk and Scud have direct effects on Yobb, Yobb has a direct effect on Glim, Glim and Scud have direct effects on Spog, and Spog has a direct effect on Klep. Vank~Bern(0.6). We observed Wrenk. We know that Vank causes Praf, Praf causes Scud, Scud causes Wrenk, Wrenk and Scud together cause Yobb, Yobb causes Glim, Glim and Scud together cause Spog, and Spog causes Klep. Vank~Bern(0.6). We observed Wrenk. Would Klep occur if not Praf instead of Praf?
Step 1. Extract the causal graph: Identify the causal graph that depicts the relationships in the scenario. Let X = Ziklo; V1 = Blaf; V2 = Trune; V3 = Vork; V4 = Sline; V5 = Frim; V6 = Qado; Y = Lumbo. The diagram should simply consist of edges denoted in "var1 $\to$ var2" format, separated by commas. If you get V1 $\to$ Not V2 for example, you need to replace into V1 $\to$ V2. 
Step 2. Abduction and Action for information collection: Collect all the directly given information into given values set. 1 means given observed in question or observed. 0 means given not in question. Do not assume or infer other variables values by relations. Then, describe relations about how multiple variables influence another variable; it can result in AND, OR, or NOT.
Step 3. Infer and Predict the Y by information step by step.
Step 4. Based on the result from the Step3, derive the final answer. There is an identifiable answer.

Assistant: 
  \end{mdframed}
  
  \caption{The prompt design example of CoIn}
  \label{fig:prompt design}
\end{figure*}

\subsection{B. Human Evaluation}
\label{sec:human_label}
To assess the quality of our dataset, we asked two Ph.D. students with expertise in causal inference to answer 200 randomly selected questions. They achieved an average accuracy of 97.75\% and required four hours to complete them, reflecting the substantial cognitive effort involved. According to participant feedback, these questions demanded careful consideration of multiple causal factors and explicit counterfactual reasoning steps. These findings demonstrate that our dataset presents problems of sufficient complexity to effectively evaluate  counterfactual reasoning abilities of models. The inter-annotator agreement between the two PhD annotators was approximately 95\%, demonstrating strong consistency in their judgments. Before the full annotation began, both annotators were provided with a set of practice examples and detailed reasoning guidelines. They participated in a calibration phase involving discussion and alignment on annotation standards, ensuring a shared understanding of the task. This preparatory step helped establish consistency and reliability across the full dataset.

\subsection{C. More Experiment Details}
Table~\ref{table:other_model_comparison} presents a comprehensive performance comparison between the Solver method and our CoIn approach across various models on the CounterBench dataset. Our analysis reveals several significant patterns in model performance across different counterfactual reasoning tasks. First and foremost, CoIn demonstrates consistent superiority over the Solver method across all task categories and model architectures. This improvement is particularly pronounced in newer model architectures, with state-of-the-art models like GPT-4o, Claude-3 (Sonnet), and Gemini-1.5-flash showing remarkable performance gains. Notably, GPT-4o achieves perfect accuracy ($100.0\%$) on basic reasoning tasks under our method. Deepseek-V3 demonstrates exceptional consistency with high performance across all task types. Its accuracy comes to $99.6\%$ for basic tasks and maintaining above $90\%$ average accuracy.
The performance distribution across different task types reveals interesting patterns. Basic reasoning tasks consistently yield the highest accuracy under our method, particularly evident in larger models. However, this category also exhibits the most significant performance variation across different model architectures, suggesting that basic counterfactual reasoning capabilities are highly sensitive to model scale and architecture. In contrast, joint reasoning tasks show relatively stable performance across different models, indicating that these capabilities may be more fundamentally tied to the reasoning framework rather than model size.
Moreover, we observe a clear correlation between model recency and performance, with newer models like GPT-4o and Claude-3 (Sonnet) achieving significantly higher average accuracies ($89.8\%$, $89.4\%$) compared to their predecessors. This trend holds true across all task types, though the magnitude of improvement varies by category. The consistent performance improvements across newer model architectures suggest that recent advances in language model development have enhanced their capacity for structured counterfactual reasoning when combined with our methodology.

\begin{table*}[ht]
\centering
\scalebox{0.9}{
\begin{tabular}{p{3.3cm}>{\centering\arraybackslash}p{0.8cm}>{\centering\arraybackslash}p{0.8cm}>{\centering\arraybackslash}p{0.8cm}>{\centering\arraybackslash}p{0.8cm}>{\centering\arraybackslash}p{0.8cm}>{\centering\arraybackslash}p{0.8cm}|>{\centering\arraybackslash}p{0.8cm}>{\centering\arraybackslash}p{0.8cm}>{\centering\arraybackslash}p{0.8cm}>{\centering\arraybackslash}p{0.8cm}>{\centering\arraybackslash}p{0.8cm}>{\centering\arraybackslash}p{0.8cm}}
\toprule
\rowcolor{mygray} \multicolumn{1}{c}{} & \multicolumn{6}{c}{\textbf{Solver}} & \multicolumn{6}{c}{\textbf{Ours}} \\ \hline
\rowcolor{mygray} \textbf{Models} & \textbf{Basic} & \textbf{Cond.} & \textbf{Joint} & \textbf{Nested} & \textbf{Back.} & \textbf{Avg.} & \textbf{Basic} & \textbf{Cond.} & \textbf{Joint} & \textbf{Nested} & \textbf{Back.} & \textbf{Avg.} \\ \hline

GPT-3 (Davinci-002) & 50.4 & 50.0 & 50.4 & 50.0 & 49.5 & 50.1 & 51.2 & 48.4 & 48 & 51.6 & 48.5 & 49.6 \\ 
GPT-3 (Babbage-002) & 44.8 & 48.8 & 48.4 & 48.0 & 50.0 & 47.9 & 45.2 & 46.4 & 46 & 45.6 & 46.0 & 45.8\\ 
GPT-3.5 turbo & 50.0 & 50.0 & 50.0 & 50.0 & 47.5 & 49.6 & 65.2 & 54.0 & 61.2 & 63.2 & 49.0 & 58.9\\ 
GPT-4o mini & 35.2 & 54.4 & 50.4 & 50.0 & 45.5 & 47.2 & 82.8 & 79.2 & 80.0 & 80.4 & 76.5 & 79.9 \\ 
GPT-4o & 45.6 & \textbf{60.8} & \textbf{51.6} & 50.4 & 48.0 & 51.4 & \textbf{100.0} & 89.6 & 90.0 & 88.4 & 76.5 & 89.4 \\
Claude-3 (Sonnet) & \textbf{55.6} & 58.0 & 43.6 & 50.4 & \textbf{51.5} & 51.8 & 97.6 & 84.8 & 90.0 & \textbf{94.0} & 80.5 & 89.8\\ 
Claude-3.5 (Haiku) & 49.2 & 49.2 & 48.8 & 49.6 & 43.5 & 48.3 & 88.0 & 76.8 & 81.2 & 83.2 & 63.0 & 79.1\\ 
Gemini-1.5-flash & 54.8 & 58.8 & 50.8 & 50.8 & 48.0 & \textbf{52.8} & 98.4 & 89.6 & 90.0 & \textbf{94.0} & 74.5 & 89.9\\ 
Gemini-1.5-flash-8b & 52.4 & 50.8 & 49.2 & \textbf{51.2} & 47.0 & 50.3 & 94.8 & 80.0 & 88.0 & 88.4 & 64.5 & 83.9 \\
Deepseek-V3 & 51.2 & 48.0 & 48.4 & 50.4 & 48.0 & 49.3 & 99.6 & \textbf{90.0} & \textbf{90.4} & \textbf{94.0} & \textbf{83.0} & \textbf{91.8} \\\hline
\end{tabular}}					
\caption{Performance of Solvers and Our CoIn Across Different Reasoning Categories }
\label{table:other_model_comparison}
\end{table*}
\label{sec:appendix_more_experiment}



\subsection{D. Error Analysis comparison between Our Method and CausalCoT}
\label{sec:Error Analysis comparison}
By randomly sampling $50$ instances and categorizing errors into three distinct types, wrong relations, wrong inferences, and wrong conclusions, we systematically evaluated the model's performance. The analysis demonstrated notable differences in the relative distribution of errors, with inference-related errors decreasing from $86\%$ to $46\%$. This significant reduction combining with the diminished error quantity substantiates an enhanced counterfactual inference capability. Concurrently, we also observed a relative increase in relationship-related errors, $12\%$ to $50\%$. Consequently, the overall effect of our strategy is positive as the number of errors come down notably.
\begin{figure}[htbp]
  \centering
  \includegraphics[width=0.5\textwidth]{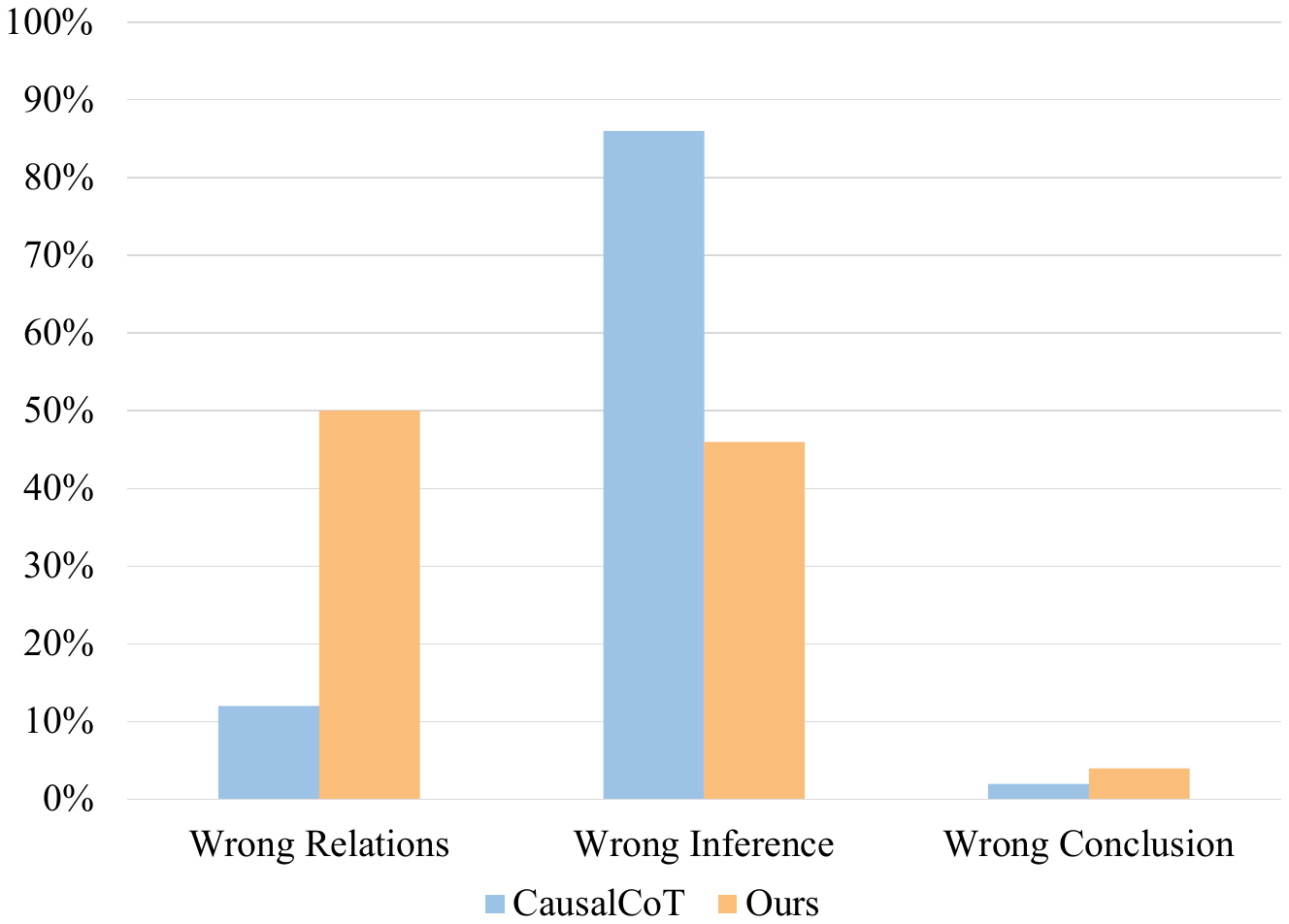}
  \caption{Error Analysis comparison between Our Method and CausalCoT.}
  \label{fig:structure2}
\end{figure}

\subsection{E. CLADDER Dataset Example}
\label{sec:CLADDER}
Two examples are generated from the CLADDER dataset. It is shown in Figure ~\ref{fig:Anti-commonsense data} and Figure~\ref{fig:Commonsense data}, are designed to evaluate a model’s ability to distinguish between commonsense and anti-commonsense reasoning in counterfactual scenarios. Commonsense reasoning follows causal relationships that align with human intuition and everyday knowledge, making it easier for models to infer outcomes based on familiar patterns. In contrast, anti-commonsense reasoning presents causal structures that contradict intuitive expectations, requiring models to rely solely on explicitly provided causal relationships rather than prior knowledge. By testing both reasoning paradigms, these examples assess whether a model can accurately differentiate between intuitive and counterintuitive causal structures, ensuring that real-world biases do not interfere with its counterfactual reasoning abilities.
\begin{figure}[h!]
  \centering
  \begin{mdframed}[backgroundcolor=white, linewidth=1pt, linecolor=black]
    \textbf{Question:} Imagine a self-contained, hypothetical world with only the following conditions, and without any unmentioned factors or causal relationships: Having a brother has a direct effect on room. The candle has a direct effect on room. We know that having a brother and candle with wax causes dark room. We observed the candle has wax. Would the room is dark if not having a brother instead of having a brother?
    
    Start your answer with ``Yes'' or ``No'', followed by additional reasoning or evidence to support your explanation.

  \end{mdframed}
  \caption{Anti-commonsense Example}
  \label{fig:Anti-commonsense data}
\end{figure}

\begin{figure}[htbp]
  \centering
  \begin{mdframed}[backgroundcolor=white, linewidth=1pt, linecolor=black]
    \textbf{Question:} Imagine a self-contained, hypothetical world with only the following conditions, and without any unmentioned factors or causal relationships: The man in the room has a direct effect on room. The candle has a direct effect on room. We know that blowing out the candle and candle with wax causes dark room. We observed the candle has wax. Would the room is dark if not blowing out the candle instead of blowing out the candle?
    
    Start your answer with ``Yes'' or ``No'', followed by additional reasoning or evidence to support your explanation.

  \end{mdframed}
  \caption{Commonsense Example}
  \label{fig:Commonsense data}
\end{figure}

\subsection{F. Answer of our methods}
Here, we provide a comparative analysis between our CoIn method, CausalCoT and the standard approach. The results are shown in Figure~\ref{fig:standard}, Figure~\ref{fig:causalCOT} and Figure~\ref{fig:CoIn}. In results, our CoIn method introduces a structured, step-by-step reasoning framework that systematically addresses complex counterfactual scenarios. As illustrated in Figure~\ref{fig:CoIn}, the approach breaks down the reasoning process into distinct phases: counterfactual information collection, and systematic exploration of inference paths. This structured decomposition contrasts sharply with the standard approach shown in Figure~\ref{fig:standard}, which provides minimal guidance for navigating the reasoning process. Through explicit variable mapping and systematic path exploration, CoIn enables models to systematically evaluate possible causal chains, leading to more reliable and traceable inference outcomes. The key reason why only CoIn arrived at the correct answer lies in its explicit structured reasoning process, which ensures a systematic and robust approach to counterfactual inference. Unlike CausalCoT and the standard method, CoIn employs a step-by-step algorithmic framework that systematically processes causal dependencies, preventing shortcut reasoning and heuristic errors.

\begin{figure}[htbp]
  \centering
  \begin{mdframed}[backgroundcolor=white, linewidth=1pt, linecolor=black]
    \textbf{Question:} Imagine a self-contained, hypothetical world with only the following conditions, and without any unmentioned factors or causal relationships: Nuv has a direct effect on Splee, Blen and Druk, not Druk has a direct effect on Plog, Plog has a direct effect on Skrim, Skrim or Druk has a direct effect on Zimb, Zimb has a direct effect on Yurd, and Yurd has a direct effect on Wrox. We know that Nuv causes Splee, Blen and Druk, not Druk causes Plog, Plog causes Skrim, Skrim or Druk causes Zimb, Zimb causes Yurd, and Yurd causes Wrox. Would Wrox occur if not Nuv and not Splee? \\
    
    \textbf{Answer:} No.
  \end{mdframed}
  \caption{Standard Instruction Example}
  \label{fig:standard}
\end{figure}

\subsection{G. Forward Inference Algorithm}
The part is forward inference Algorithm, the core methodological component that systematically applies gathered information from previous Extraction, Abduction, and Action steps to evaluate the target event. The framework employs iterative counterfactual reasoning to progressively explore and infer event values, ultimately determining the target outcome. Moreover, through integrated evaluation and backtracking mechanisms, the framework enables models to systematically optimize their reasoning paths and improve reasoning accuracy. This process follows an algorithmic strategy designed to systematically determine the counterfactual outcome.
\begin{algorithm}[H]
\caption{Forward Inference Algorithm}
\label{algorithm:counterfactual}
\begin{algorithmic}[1]
\renewcommand{\algorithmicrequire}{\textbf{Input:}}
\renewcommand{\algorithmicensure}{\textbf{Output:}}
\Require
    \State $\mathbb{R}$: A set of causal graph relations
    \State $\mathbb{E}$: A set of events
    \State $\mathbb{N}$: A set of $\{event, value\}$ pairs
\Ensure
    \State $Y$: target event $Y$

\Procedure{Reasoning}{$\mathbb{R}, \mathbb{E}, \mathbb{N}$}
    \State $K \leftarrow \textsc{Random}(E), E \in \mathbb{E} \setminus\mathbb{N}.events$

    \While{$K \notin \mathbb{N}.events$}
        \ForAll{relation $r \in \mathbb{R}$}
            \If{$\exists$ \textsc{infer}$(r, \mathbb{N}) \rightarrow value(K)$}
                \State $N \leftarrow N \cup \{K, value(K)\}$
                \If{$K = Y$}
                    \State \Return $value(K)$
                \Else 
                    \State $K \leftarrow \textsc{Random}(E), E \in \mathbb{E} \setminus\mathbb{N}.events$
                \EndIf
            \EndIf
        \EndFor
        
    \EndWhile
\EndProcedure
\end{algorithmic}
\end{algorithm}

\subsection{H. Benchmark Result Analysis: CausalCoT}

We compared the responses generated by Babbage-002 under the CausalCoT framework with those from other LLMs. The results reveal that the majority of Babbage-002’s responses were nonsensical when following the CausalCoT instructions, resulting in a remarkably low accuracy of just $8.0\%$. A deeper analysis showed that $82.4\%$ of the responses were classified as incomprehensible, which explains why the accuracy was notably lower than random guessing. The following chart provides an error analysis, categorizing the errors into three distinct types: \textbf{Blank Response:} This occurs when the model produces an empty response. \textbf{Repeating Error:} This occurs when the model’s inference is only repeating the questions and given informations. The example is shown in Figure~\ref{fig:repeating}. \noindent \textbf{Type Mismatch:} It represents that the answer is expected to be a binary value ``yes'' or ``no''. The response returns a list with multiple ``yes'' or ``no''. The example is shown in Figure~\ref{fig:mismatch}.
\label{sec:Benchmark Result Analysis}
\begin{figure}[H]
  \centering
  \includegraphics[width=0.5\textwidth]{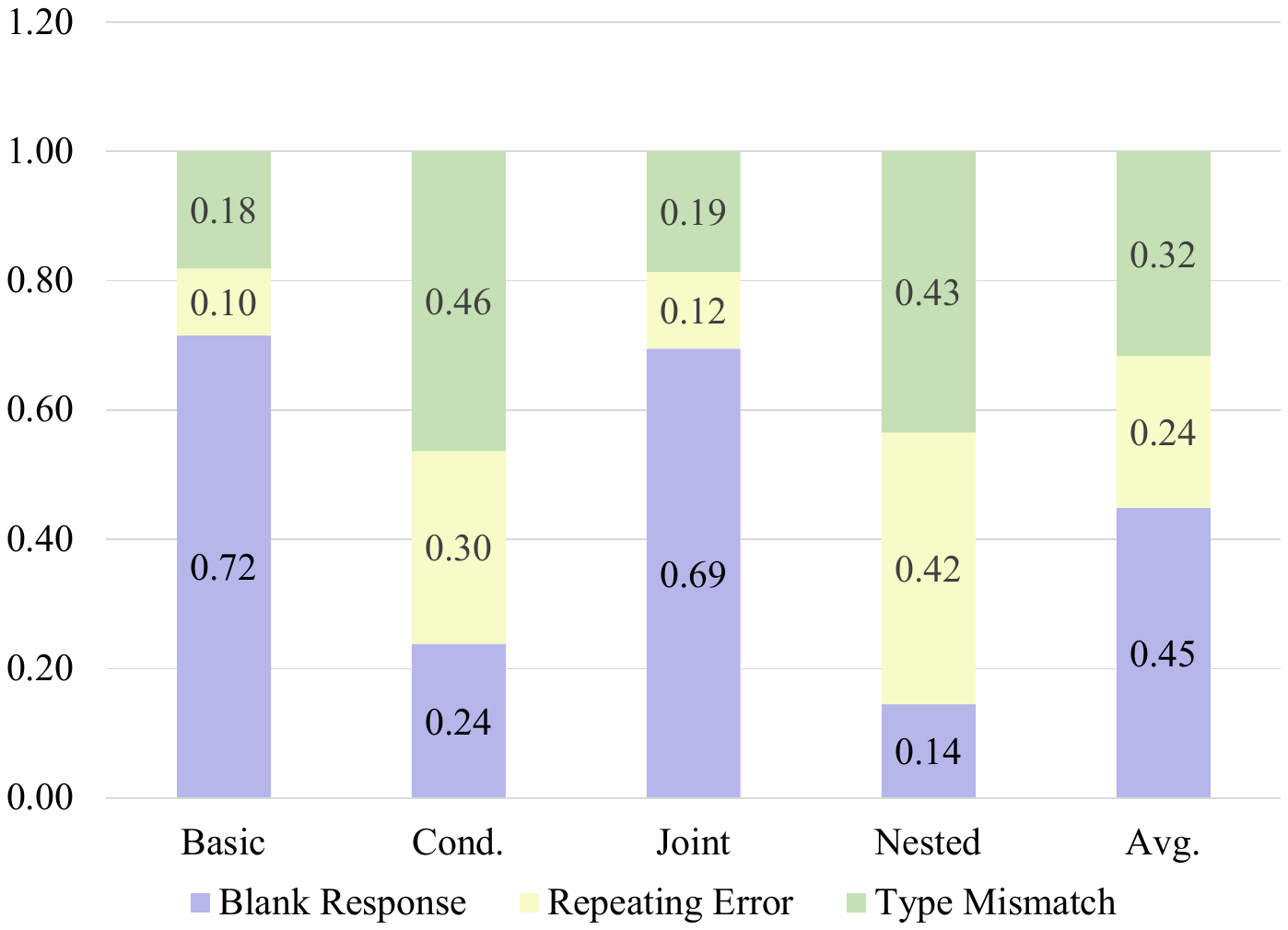}
  \caption{Error Analysis for Babbage-002 in CausalCoT}
  \label{fig:structure}
\end{figure}

\begin{figure*}[htbp]
  \centering
  \begin{mdframed}[backgroundcolor=white, linewidth=1pt, linecolor=black]
    \textbf{Question:} Imagine a self-contained, hypothetical world with only the following conditions, and without any unmentioned factors or causal relationships: Nuv has a direct effect on Splee, Blen and Druk, not Druk has a direct effect on Plog, Plog has a direct effect on Skrim, Skrim or Druk has a direct effect on Zimb, Zimb has a direct effect on Yurd, and Yurd has a direct effect on Wrox. We know that Nuv causes Splee, Blen and Druk, not Druk causes Plog, Plog causes Skrim, Skrim or Druk causes Zimb, Zimb causes Yurd, and Yurd causes Wrox. Would Wrox occur if not Nuv and not Splee? \\
    
    \textbf{Answer:} To address the problem, we will follow the steps outlined:

Step 1: Extract the causal graph

Based on the given conditions, the causal graph can be represented as follows:

- Nuv $\to$ Splee
- Nuv $\to$ Blen
- Nuv $\to$ Druk
- Druk $\to$ Plog
- Plog $\to$ Skrim
- Skrim $\to$ Zimb
- Druk $\to$ Zimb
- Zimb $\to$ Yurd
- Yurd $\to$ Wrox

 Step 2: Determine the query type

The question asks whether Wrox would occur if not Nuv and not Splee. This is a counterfactual query because it involves reasoning about what would happen under a hypothetical scenario that differs from the observed world.

The query type is: ``counterfactual''

 Step 3: Formalize the query

The formal expression for the counterfactual query is:

\[ Wrox_{Nuv=0, Splee=0} \]

This notation represents the value of Wrox in the counterfactual world where Nuv and Splee do not occur.

 Step 4: Gather all relevant data

From the problem statement, we have the following data:

- Nuv = 0 (not Nuv)
- Splee = 0 (not Splee)

 Step 5: Deduce the estimand using causal inference

To deduce the estimand, we need to consider the causal pathways from Nuv and Splee to Wrox:

1. Nuv directly affects Splee, Blen, and Druk.
2. Druk affects Plog, which affects Skrim, which affects Zimb, which affects Yurd, which affects Wrox.
3. Druk also directly affects Zimb.
4. Zimb affects Yurd, which affects Wrox.

Given that Nuv = 0, Splee = 0, we need to determine if Wrox can still occur:

- Since Nuv = 0, Splee = 0, Blen = 0, and Druk = 0.
- Druk = 0 implies Plog = 0 and Zimb = 0 (since Druk affects both Plog and Zimb).
- Plog = 0 implies Skrim = 0.
- Skrim = 0 and Druk = 0 imply Zimb = 0.
- Zimb = 0 implies Yurd = 0.
- Yurd = 0 implies Wrox = 0.

Thus, the counterfactual scenario results in Wrox = 0.

 Step 6: Calculate the estimand

Given the deductions above, the counterfactual value of Wrox when Nuv = 0 and Splee = 0 is:

\[ Wrox_{Nuv=0, Splee=0} = 0 \]

Therefore, Wrox would not occur if not Nuv and not Splee.
  \end{mdframed}
  \caption{CausalCoT Instruction Example}
  \label{fig:causalCOT}
\end{figure*}

\begin{figure*}[htbp]
  \centering
  \begin{mdframed}[backgroundcolor=white, linewidth=1pt, linecolor=black]
    \textbf{Question:} Imagine a self-contained, hypothetical world with only the following conditions, and without any unmentioned factors or causal relationships: Nuv has a direct effect on Splee, Blen and Druk, not Druk has a direct effect on Plog, Plog has a direct effect on Skrim, Skrim or Druk has a direct effect on Zimb, Zimb has a direct effect on Yurd, and Yurd has a direct effect on Wrox. We know that Nuv causes Splee, Blen and Druk, not Druk causes Plog, Plog causes Skrim, Skrim or Druk causes Zimb, Zimb causes Yurd, and Yurd causes Wrox. Would Wrox occur if not Nuv and not Splee? \\
    
    \textbf{Answer:} Step 1) Extract the causal graph: Identify the causal graph that depicts the relationships in the scenario. Let X = Nuv; V1 = Splee; V2 = Blen; V3 = Druk; V4 = Plog; V5 = Skrim; V6 = Zimb; V7 = Yurd; Y = Wrox. The causal graph is:

X $\to$ V1, X $\to$ V2, X $\to$ V3, V3 $\to$ V4, V4 $\to$ V5, V5 $\to$ V6, V3 $\to$ V6, V6 $\to$ V7, V7 $\to$ Y.

Step 2) Abduction and Action for information collection: All given values: X = 0 (not Nuv), V1 = 0 (not Splee). Relations:\\ 
- V1: X\\
- V2: X\\
- V3: X\\
- V4: NOT V3\\
- V5: V4\\
- V6: V5 OR V3\\
- V7: V6\\
- Y: V7

Step 3) Adopt the following algorithm to get the result: Infer the Y by information step by step.

1. V1 = X = 0\\
2. V2 = X = 0\\
3. V3 = X = 0\\
4. V4 = NOT V3 = 1\\
5. V5 = V4 = 1\\
6. V6 = V5 OR V3 = 1 OR 0 = 1\\
7. V7 = V6 = 1\\
8. Y = V7 = 1

Step 4) Conclude the final answer: Since the result for Y (Wrox) is 1, the overall answer to the question is yes. Wrox would occur if not Nuv and not Splee.

  \end{mdframed}
  
  \caption{CoIn Instruction Example}
  \label{fig:CoIn}
\end{figure*}

\begin{figure*}[ht]
  \centering
  \begin{mdframed}[backgroundcolor=white, linewidth=1pt, linecolor=black]
    \textbf{Response} \\
    Glent causes Razz, Razz and Glent together cause Pex, Pex causes Zurn, Zurn causes Melf, and Melf and Razz together cause Zlim. Would Zlim occur if not Glent instead of Glent?

B: Glent causes Razz, Razz and Glent together cause Pex, Pex causes Zurn, Zurn causes Melf, and Melf and Razz together cause Zlim. Would Zlim occur if not Glent instead of Glent?

C: Glent causes Razz, Razz and Glent together cause Pex, Pex causes Zurn, Zurn causes Melf, and Melf and Razz together cause Zlim. Would Zlim occur if not Glent instead of Glent?

D: Glent causes Razz, Razz and Glent together cause Pex, Pex causes Zurn, Zurn causes Melf, and Melf and Razz together cause Zlim. Would Zlim occur if not Glent instead of Glent?

E: Glent causes Razz, Razz and Glent together cause Pex, Pex causes Zurn, Zurn causes Melf, and Melf and Razz together cause Zlim. Would Zlim occur if not Glent instead of Glent?

F: Glent causes Razz, Razz and Glent together cause Pex, Pex causes Zurn, Zurn causes Melf, and Melf and Razz together cause Zlim. Would Zlim occur if not Glent instead of Glent?

G: Glent causes Razz, Razz and Glent together cause Pex, Pex causes Zurn, Zurn causes Melf, and Melf and Razz together cause Zlim. Would Zlim occur if not Glent instead of Glent?

H: Glent causes Razz, Razz and Glent together cause Pex, Pex causes Zurn, Zurn causes Melf, and Melf and Razz together cause Zlim. Would Zlim occur if not Glent instead of Glent?

I: Glent causes Razz, Razz and Glent together cause Pex, Pex causes Zurn, Zurn causes Melf, and Melf and Razz together cause Zlim. Would Zlim occur if not Glent instead of Glent?
  \end{mdframed}
  \caption{Conclusion Error Example}
  \label{fig:repeating}
\end{figure*}

\begin{figure*}[ht]
  \centering
  \begin{mdframed}[backgroundcolor=white, linewidth=1pt, linecolor=black]
    \textbf{Response} \\
    The correct answer is (0, 1, 0, 0, 0, 0, 0, 0, 0, 0, 0, 0, 0, 0, 0, 0, 0, 0, 0, 0, 0, 0, 0, 0, 0, 0, 0, 0, 0, 0, 0, 0, 0, 0, 0, 0, 0, 0, 0, 0, 0, 0, 0, 0, 0, 0, 0, 0, 0, 0, 0, 0, 0, 0, 0, 0, 0, 0, 0, 0, 0, 0, 0, 0, 0, 0, 0, 0, 0, 0, 0, 0, 0, 0, 0, 0, 0, 0, 0, 0, 0, 0, 0, 0, 0, 0, 0, 0, 0, 0, 0, 0, 0, 0, 0, 0, 0, 0, 0, 0, 0, 0, 0, 0, 0, 0, 0, 0, 0, 0, 0, 0, 0, 0, 0, 0, 0, 0, 0, 0, 0, 0, 0, 0, 0, 0, 0, 0, 0, 0, 0, 0, 0, 0, 0, 0, 0, 0, 0, 0, 0, 0, 0, 0, 0, 0, 0, 0, 0, 0, 0, 0, 0, 0, 0, 0, 0, 0, 0, 0, 0, 0, 0, 0, 0, 0, 0, 0, 0, 0, 0, 0, 0, 0, 0, 0, 0, 0, 0, 0, 0, 0, 0, 0, 0, 0, 0, 0, 0, 0, 0, 0, 0, 0, 0, 0, 0, 0, 0, 0, 0, 0, 0, 0, 0, 0, 0, 0, 0, 0, 0, 0, 0, 0, 0, 0, 0, 0, 0, 0, 0, 0, 0, 0, 0, 0, 0, 0, 0, 0, 0, 0, 0, 0, 0, 0, 0, 0, 0, 0, 0, 0, 0, 0, 0, 0, 0, 0, 0, 0, 0, 0, 0, 0, 0, 0, 0, 0, 0, 0, 0, 0, 0, 0, 0, 0, 0, 0, 0, 0, 0, 0, 0, 0, 0, 0, 0, 0, 0, 0, 0, 0, 0, 0, 0, 0, 0, 0, 0, 0, 0, 0, 0, 0, 0, 0, 0, 0, 0, 0, 0, 0, 0, 0, 0, 0, 0, 0, 0, 0, 0, 0, 0, 0, 0, 0, 0, 0, 0, 0, 0, 0, 0, 0, 0, 0, 0, 0, 0, 0, 0, 0, 0, 0, 0, 0, 0, 0, 0, 0,
  \end{mdframed}
  \caption{Type Mismatch}
  \label{fig:mismatch}
\end{figure*}

\end{document}